\documentclass[final, conference]{IEEEtran}
\IEEEoverridecommandlockouts
\usepackage{cite}
\usepackage{amsmath,amssymb,amsfonts}
\usepackage[ruled,noend,linesnumbered]{algorithm2e}
\usepackage{booktabs}
\usepackage{tabularx}
\usepackage{hyperref}
\makeatletter
\newcommand{\removelatexerror}{\let\@latex@error\@gobble}
\makeatother
\usepackage{graphicx}
\usepackage{textcomp}
\usepackage{xcolor}
\usepackage{subfigure}
\usepackage{hyperref}
\usepackage{cleveref}
\usepackage{caption}
\usepackage[all]{hypcap}
\usepackage{xcolor}
\hypersetup{
    colorlinks,
    linkcolor={red!50!black},
    citecolor={blue!50!black},
    urlcolor={blue!80!black}
}

\usepackage[colorinlistoftodos]{todonotes}
\usepackage{xcolor}
\usepackage[most]{tcolorbox}
\usepackage{todonotes}
\usepackage{color,soul}
\usepackage{tikz}
\usepackage[switch]{lineno}

\usepackage{scalerel}
\usetikzlibrary{svg.path}

\definecolor{orcidlogocol}{HTML}{A6CE39}
\tikzset{
  orcidlogo/.pic={
    \fill[orcidlogocol] svg{M256,128c0,70.7-57.3,128-128,128C57.3,256,0,198.7,0,128C0,57.3,57.3,0,128,0C198.7,0,256,57.3,256,128z};
    \fill[white] svg{M86.3,186.2H70.9V79.1h15.4v48.4V186.2z}
                 svg{M108.9,79.1h41.6c39.6,0,57,28.3,57,53.6c0,27.5-21.5,53.6-56.8,53.6h-41.8V79.1z M124.3,172.4h24.5c34.9,0,42.9-26.5,42.9-39.7c0-21.5-13.7-39.7-43.7-39.7h-23.7V172.4z}
                 svg{M88.7,56.8c0,5.5-4.5,10.1-10.1,10.1c-5.6,0-10.1-4.6-10.1-10.1c0-5.6,4.5-10.1,10.1-10.1C84.2,46.7,88.7,51.3,88.7,56.8z};
  }
}

\newcommand\orcidicon[1]{\href{https://orcid.org/#1}{\mbox{\scalerel*{
\begin{tikzpicture}[yscale=-1,transform shape]
\pic{orcidlogo};
\end{tikzpicture}
}{|}}}}

\NewDocumentCommand \T { O{} m } {\ensuremath{\boldsymbol{#1\mathscr{\MakeUppercase{#2}}}}}
\NewDocumentCommand \M { O{} m } {\ensuremath{\bm{#1\mathbf{\MakeUppercase{#2}}}}} 
\NewDocumentCommand \V { O{} m } {\ensuremath{\bm{#1\mathbf{\MakeLowercase{#2}}}}} 

\definecolor{anzheng}{RGB}{255,165,0}

\def\tablename{Table}
\def\BibTeX{{\rm B\kern-.05em{\sc i\kern-.025em b}\kern-.08em
    T\kern-.1667em\lower.7ex\hbox{E}\kern-.125emX}}
\begin{document}

\title{Im2win: Memory Efficient Convolution On SIMD Architectures}

\author{\IEEEauthorblockN{Shuai Lu}
\IEEEauthorblockA{\textit{Nanchang Hangkong University}\\
Nanchang, China \\
2016085400101@stu.nchu.edu.cn}
\and
\IEEEauthorblockN{Jun Chu}
\IEEEauthorblockA{\textit{Nanchang Hangkong University}\\
Nanchang, China \\
chuj@nchu.edu.cn}
\and
\IEEEauthorblockN{Xu T. Liu~\orcidicon{0000-0003-3980-9803}}
\IEEEauthorblockA{\textit{University of Washington}\\
Seattle, USA \\
x0@uw.edu}
}

\maketitle

\thispagestyle{plain}
\pagestyle{plain}

\begin{abstract}
Convolution is the most expensive operation among neural network operations, thus its performance is critical to the overall performance of neural networks. Commonly used convolution approaches, including general matrix multiplication (GEMM)-based convolution and direct convolution, rely on im2col for data transformation or do not use data transformation at all, respectively. However, the im2col data transformation can lead to at least 2$\times$ memory footprint compared to not using data transformation at all, thus limiting the size of neural network models running on memory-limited systems. Meanwhile, not using data transformation usually performs poorly due to nonconsecutive memory access although it consumes less memory. To solve those problems, we propose a new memory-efficient data transformation algorithm, called im2win. This algorithm refactorizes a row of square or rectangle dot product windows of the input image and flattens unique elements within these windows into a row in the output tensor, which enables consecutive memory access and data reuse, and thus greatly reduces the memory overhead. Furthermore, we propose a high-performance im2win-based convolution algorithm with various optimizations, including vectorization, loop reordering, etc. 
Our experimental results show that our algorithm reduces the memory overhead by average to 41.6\% compared to the PyTorch's convolution implementation based on im2col, and achieves average to 3.6$\times$ and 5.3$\times$ speedup in performance compared to the im2col-based convolution and not using data transformation, respectively.
\end{abstract}


\section{Introduction}

Convolution is one of the most important components within neural network models, not only because it helps filter critical features out of massive data efficiently, but also because it is the most expensive operation compared with other operations~\cite{crowley2018moonshine}. 
In particular, a recent study~\cite{shufflenet} reported that, in CNN inference, there are 50\%-90\% of the total operations, including pooling, ReLU, and fully-connected, that are convolution operations.
Also, convolution accounts for over 90\% of the total execution time of many popular neural networks~\cite{efficient_processing_of_DNN}.
Therefore, it is essential to reduce the cost of convolution operations in order to improve performance of neural networks. 

\begin{figure}[tp]
\centering
\centerline{\includegraphics[scale=0.45]{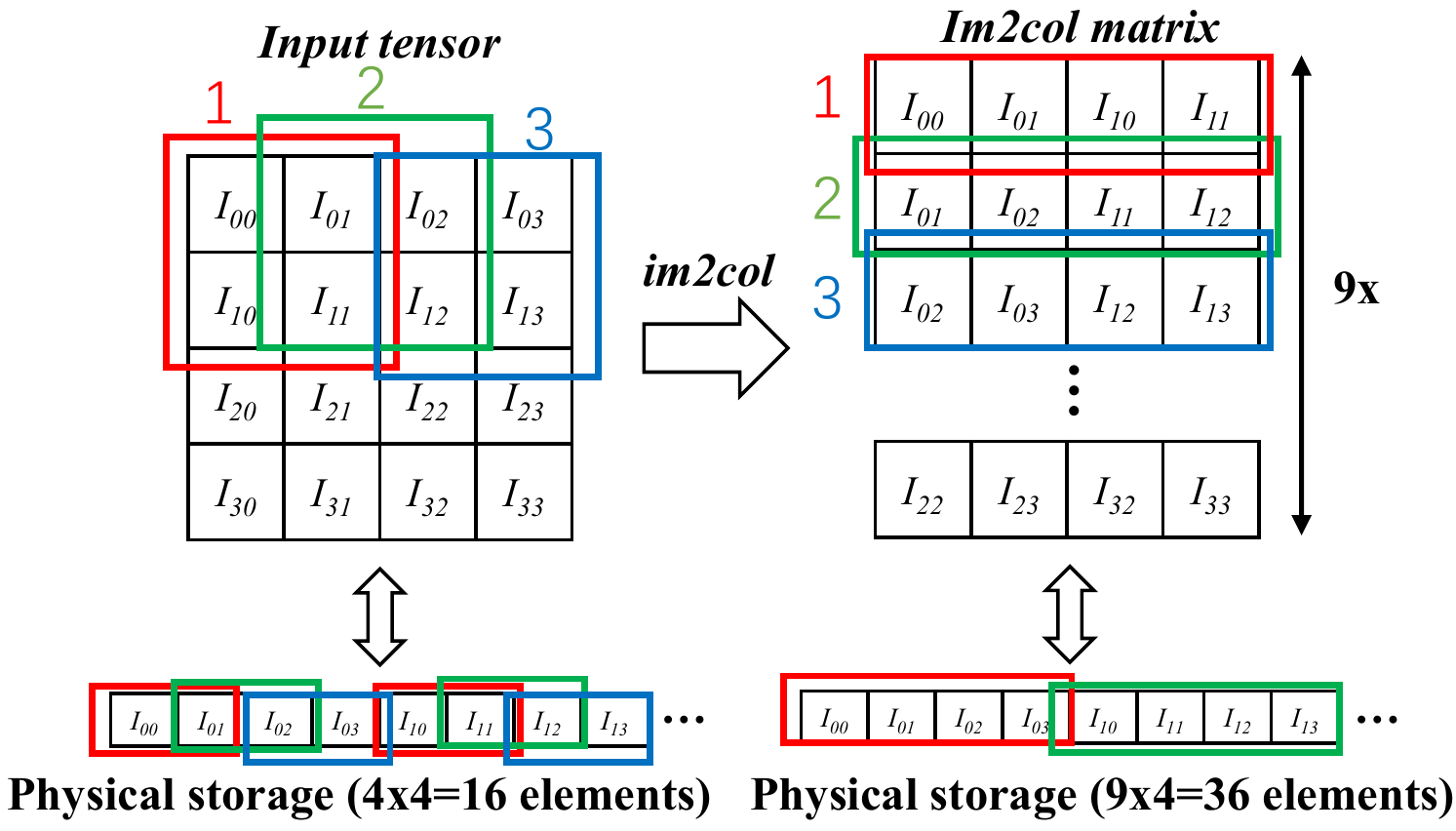}}
\caption{\small Illustration of the im2col transformation on a 1$\times$1$\times$4$\times$4 input tensor and a 1$\times$1$\times$2$\times$2 filter tensor. The red, green and blue boxes indicate the window for the first three dot product operations. Note after transformation, im2col matrix has 36 elements.}
\label{fig:im2col_illustration}
\end{figure}

There are mainly two groups of convolution approaches with respect to data transformation within production machine learning frameworks, such as PyTorch~\cite{pytorch} and TensorFlow~\cite{tensorflow}: im2col data transformation-based and not using data transformation at all, also known as direct convolution. 
Im2col data transformation takes elements of each dot product window, flattens them out and stacks them up row by row in the output matrix, called im2col matrix. Im2col-based convolution operation starts by transforming the input tensor into im2col matrix by im2col algorithm, and the filter is unfolded into a 2D matrix, 
followed by a GEMM function call using Basic Linear Algebra Subprograms (BLAS)~\cite{blas} to complete the compute operation, and finally transposes the resultant matrix into a high-dimensional tensor. The problem with the im2col-based convolution is that the im2col operation generates a high memory footprint and bandwidth overhead. As shown in~\Cref{fig:im2col_illustration}, the im2col transformation produces a matrix with many duplicate elements, for example, the elements in the green box are stored twice in the im2col matrix. 
In the rest of the paper, we assume all data (such as images, matrices, and tensors) are stored as an 1D array in memory.
The resultant im2col matrix is typically larger than the input image (36 elements versus 16 elements). 

Convolution approaches without data transformation often use direct convolution algorithm~\cite{directconvolutions}. 
A typical direct convolution is implemented with seven nested loops over the original input tensor with scalar alpha times x plus y (AXPY) computations.
This method has no additional memory overhead compared to im2col-based (GEMM) convolution,  
however, its memory access is nonconsecutive.
As shown on the left in~\Cref{fig:im2col_illustration},
each dot product involving the window data is nonconsecutive memory access, resulting low data reuse and cache hit. The larger the input tensor is, the more the performance of direct convolution degrades.



To solve those problems, we propose a new memory-efficient data transformation algorithm, called the im2win algorithm. 
In contrast with im2col, our im2win algorithm refactorizes a row of square or rectangle dot product windows of the input image and flattens unique elements within these windows into a row in the output tensor, which enables consecutive memory access and data reuse, and thus greatly reduces the memory overhead. Note that there may be duplicate elements between row and row.
Furthermore, we propose a high-performance im2win-based convolution algorithm with various optimizations, including vectorization and fused-multiply-add (FMA) instructions, loop reordering,
hoist and loop unrolling, register and cache blocking, and parallelization strategy.
Our experimental results show that our algorithm reduces the memory overhead by average to 41.6\% compared to the PyTorch's up-to-date convolution implementation based on im2col, and achieves on average 3.6$\times$ and 5.3$\times$ speedup in performance compared to the im2col-based convolution and not using data transformation, respectively. We make our code publicly available at \href{https://github.com/ls110082/Im2win}{https://github.com/ls110082/Im2win}.

\noindent\textbf{Contributions.} The main contributions of this paper are:\\
\indent
1) We propose an~\textbf{im2win data transformation algorithm}, which not only greatly reduces the memory footprint for convolutions, but also provides better data locality. This contributes to our overall 41.6\% memory reduce against PyTorch's im2col-based convolution. 
\\
\indent 2) We propose an~\textbf{im2win-based convolution algorithm} which makes the best use of the data reuse and locality provided by our im2win data transformation algorithm.
\\
\indent 3) We propose a collection of optimizations for~\textbf{the im2win-based convolution algorithm} on SIMD architecture with  vectorization instruction set and various optimizations.

\section{Preliminaries}
In this section, we define the notations used in this paper, review the im2col-based convolution and direct convolution, and the related works.
\subsection{Notations}

The three main tensor data in the convolution operation are 
the Input tensor ($\mathcal{I}$), the Filter tensor ($\mathcal{F}$), and the Output tensor ($\mathcal{O}$). 
These tensors in NCHW layout are expressed as $\mathcal{I}[N_{i}][C_{i}][H_{i}][W_{i}]$, $\mathcal{F}[C_{o}][C_{i}][H_{f}][W_{f}]$ and $\mathcal{O}[N_{i}][C_{o}][H_{o}][W_{o}]$. The convolution is defined as:
\begin{equation}
\begin{array}{r}
\begin{aligned}
\mathcal{O}_{(i, j, m, n)}=\sum_{j=0}^{C_{i}-1} \sum_{m=0}^{H_{f}-1} \sum_{n=0}^{W_{f}-1}\left(\mathcal{I}_{(i, j, m \times s+u, m \times s+v)}\right. \\
\left.\times \mathcal{F}_{(j, r, u, v)}\right),
\end{aligned}
\end{array}
\end{equation}
subject to
\begin{equation}
\begin{aligned}
&i~=0,1,..,N_{i}-1, j=0,1,..,C_{o}-1, m=0,1,..,H_{o}-1,\\
&n=0,1,..,W_{o}-1, u=0,1,..,H_{f}-1, v=0,1,..,W_{f}-1,\\
&r~=0,1,..,C_{i}-1. \nonumber
\end{aligned}
\end{equation}
$N_{i}$ is the batch size, $s$ is the stride size, $C_{i}$ and $C_{o}$ are the number of input and output channels, $H_{i/f/o}$ and $W_{i/f/o}$ denote height and width in spatial dimensions.


\subsection{The im2col-based Convolution}

\begin{figure}[htbp]
\centering
\centerline{\includegraphics[scale=0.45]{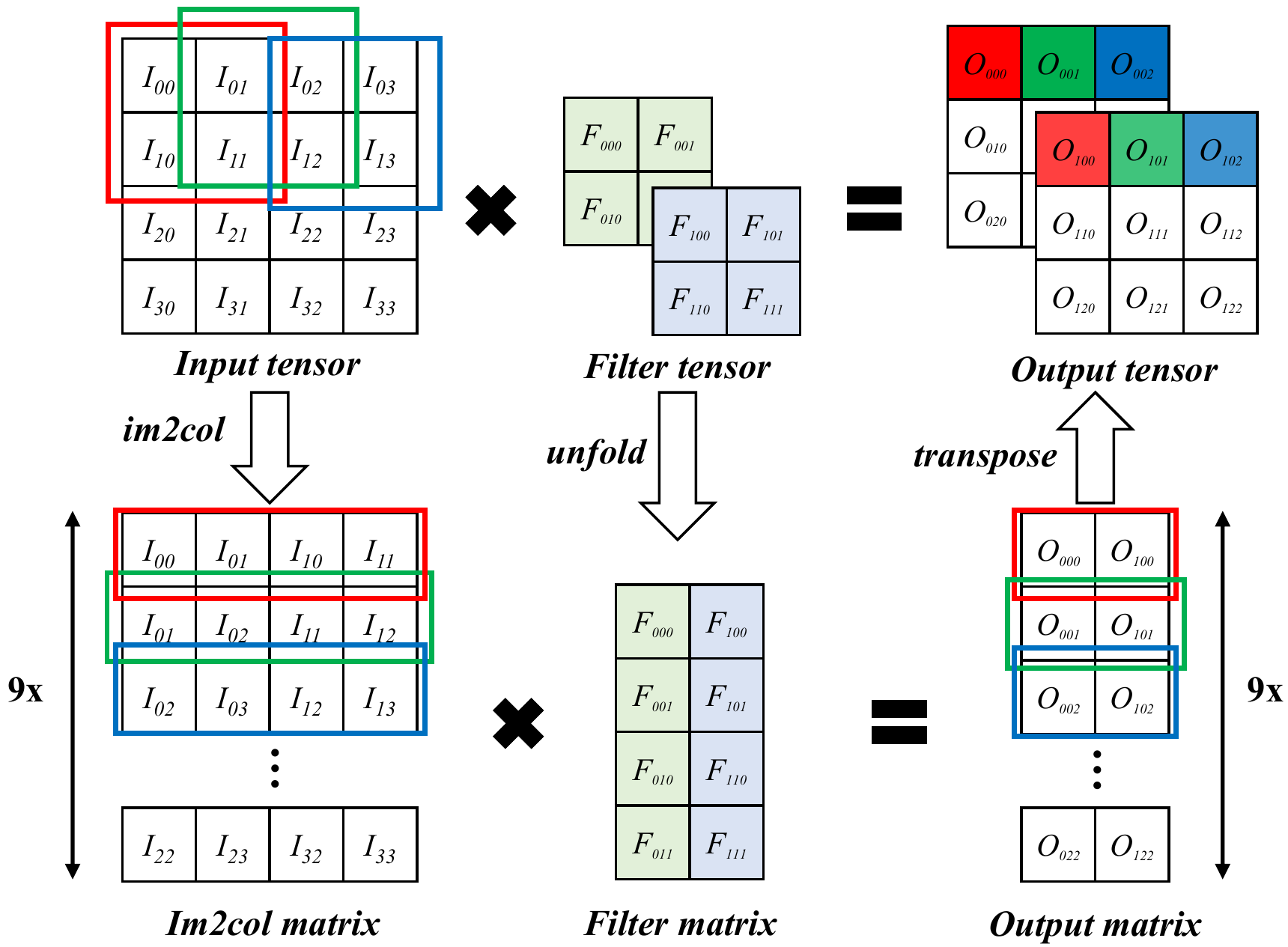}}
\caption{\small Basic convolution and im2col+GEMM convolution examples with $H_{i}$ = $W_{i}$ = 4, $H_{f}$ = $W_{f}$ = 2, $s_{h}$ = $s_{w}$ = 1, $H_{o}$ = $W_{o}$ = 3 ($C_{i}$ = $C_{f}$ = 1, $C_{o}$ = 2). The different colored boxes denote the correspondence between the input image and the output features.}
\label{fig:im2col+GEMM_convolution}
\end{figure}
Proposed by Chellapilla et al.~\cite{chellapilla_high_2006}, 
the im2col-based convolution is the simplest and most commonly used convolution algorithm, and it is widely used in existing deep learning frameworks~\cite{pytorch,tensorflow,caffe}. Due to its fundamental and general nature, it is often used as a benchmark for comparison.
The im2col-based convolution unrolls the convolution operation into a general matrix multiplication operation. The $\mathcal{I}$ of size $N_i \times C_i \times H_i \times W_i$ is processed in $N_i$ batches, each batch contains data $\mathcal{I'}$ of size $C_i \times H_i \times W_i$ (i.e., a single image). As shown in~\Cref{fig:im2col+GEMM_convolution}, the im2col algorithm 
transforms $\mathcal{I'}$ into a 2D matrix; and $\mathcal{F}$ is unfolded into a filter matrix. 
In im2col, the elements of each dot product window of $\mathcal{I'}$ is flattened and copied into a single row of a matrix (see~\Cref{fig:im2col_illustration}).
Denoting the im2col matrix as $M$ and the filter matrix as $N$, the im2col algorithm can be written as: $M(m W_{o}+n, (r H_{f}+u)W_{f}+v)=\mathcal{I'}(r, m+u, n+v),~N((r H_{f}+u)W_{f}+v, j)=\mathcal{F}(j, r, u, v)$. 
Next, a GEMM operation in BLAS library performs the matrix product of the transformed input matrix and the transformed filter matrix to get the output matrix: $R'=M\times N$.
The convolution result tensor $R$ is transposed from $R'$: $R(j,m,n)=R'(m W_{o}+n,j)$.

Benefiting from the highly optimized BLAS library, the im2col-based convolution has reliable performance and supports various input tensor sizes~\cite{chellapilla_high_2006}. 
However, the performance of this convolution depends heavily on the performance of the GEMM operation.
Previous work finds that when the input matrices vary significantly in size and shape, the GEMM operation performs poorly on hierarchical memory architectures~\cite{a_family_of_MM}.
Because the im2col matrix is much larger than the filter matrix, this results the GEMM operation in significantly lower performance than the best achievable performance.
Several im2col-based works~\cite{Low-memorygemm,p-im2col} and MEC~\cite{mec} split a single matrix multiplication into multiple small matrix multiplications to reduce the storage overhead. MEC proposes a matrix lowering scheme for the input matrix to reduce the memory footprint, then performs multiple GEMM operations in parallel to complete the convolution.
Different from these approaches, we do not perform matrix multiplication in our convolution, instead, we transform the input tensor and perform convolution on the transformed tensor.


\subsection{Direct Convolution}


The direct convolution process is described in the top of~\Cref{fig:im2col+GEMM_convolution}. 
It performs direct convolution on
$\mathcal{I}$ and $\mathcal{F}$ without transformation,
hence it has low memory and bandwidth usage compared with the im2col-based convolution. A basic direct convolution has seven nested for loops. The outer four loops iterate over the four dimensions of $\mathcal{O}$, and the inner three loops iterate over $\mathcal{F}$ and $\mathcal{I}$. The result is obtained by performing a basic AXPY operation within the above loops.
Direct convolution costs no additional memory but accesses data nonconsecutively therefore has low performance.


Some recent works in high performance computing have focused on direct convolution. Intel LibXSMM library employs parameterized architecture-specific Just-in-Time code generator to optimize the implementation of direct convolution~\cite{ibxsmm}.
It has been shown that the performance of direct convolution can be greatly improved by designing specific data layouts based on the loop ordering of the algorithm~\cite{directconvolutions}. The optimizations on SIMD architecture have also been proposed~\cite{direct_conv_on_SIMD}.
Different from the above works, our work
provides consecutive memory access and reduces additional memory cost by making the best use of the proposed im2win data transformation.



\section{Our Im2win-based Convolution}




To solve the huge memory footprint of the im2col-based convolution, and the nonconsecutive memory access of the direct convolution, 
we present our im2win data transformation algorithm and our high-performance im2win-based convolution algorithm. 
In addition, we will elaborate on the implementation of our high-performance im2win-based convolution algorithm and the optimization techniques.

\begin{figure}[tbp]
\centering
\centerline{\includegraphics[scale=0.45]{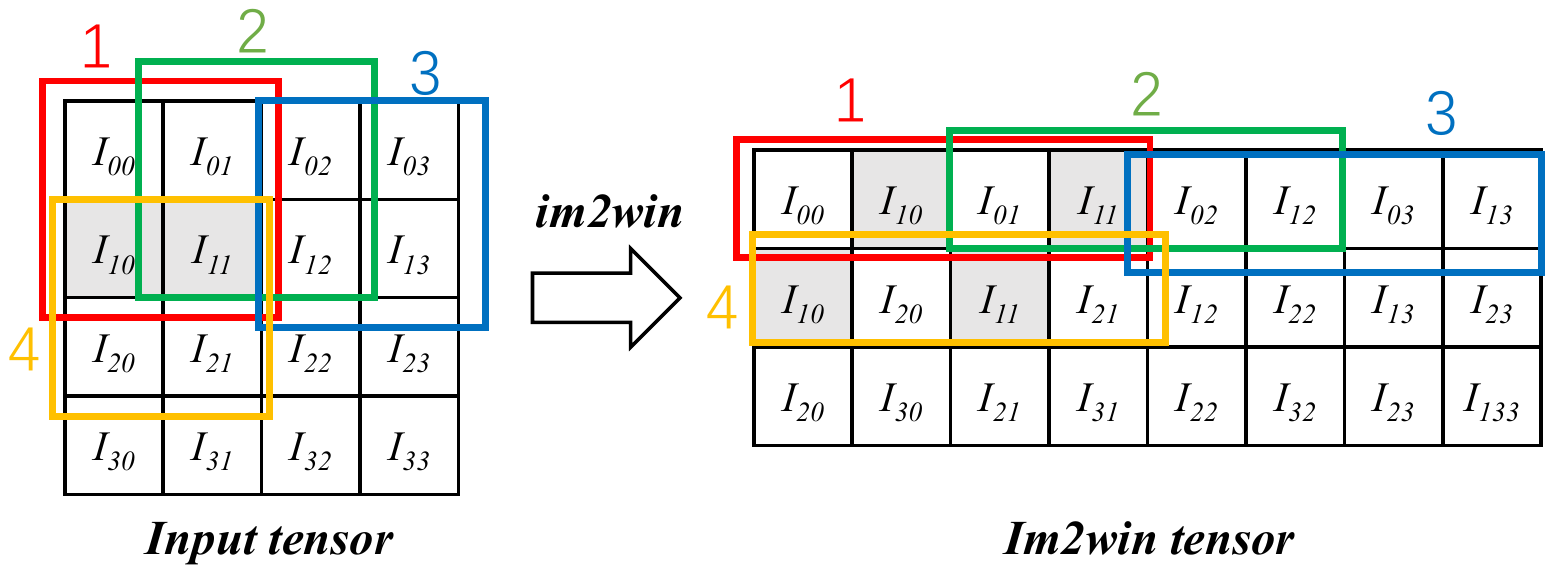}}
\caption{\small
Illustration of the im2win data transformation from the original tensor to the im2win tensor.
The different colored boxes indicate the mapping of elements between the input and the im2win tensors. After transformation, the im2win tensor has 24 elements.}
\label{fig:im2win_illustration}
\end{figure}

\subsection{Proposed Im2win Algorithm}\label{subsec:im2win}
We propose an image to window algorithm (called im2win), which dramatically reduce memory overhead 
by more compact data arrangement compared with im2col. 
As shown in~\Cref{fig:im2win_illustration}, our algorithm transforms $\mathcal{I}$ in the order of the dot product window in the convolution operation. For two consecutive dot product window operations, most of the elements loaded in the first operation are reused in the second operation, which can greatly improve data reusability and cache hit.

In the im2win algorithm, we divide each channel of $\mathcal{I'}$ into $H_o\times W_o$ windows of size $H_f\times W_f$, and copy $W_o$ windows in the same row to one row in our im2win tensor. 
Performing the above operation for all windows on a single channel of $\mathcal{I'}$, we obtain a tensor of size ($H_o $, $H_f \times W_i$) (see~\Cref{fig:im2win_illustration}). This tensor is ordered by the dot product windows and has fewer redundant elements than what the im2col matrix has. Performing the above algorithm for the batch and channel dimensions in $\mathcal{I}$, we will get a tensor of size ($N_i$, $C_i$, $H_o $, $W_i \times H_f$) and call this tensor as an im2win tensor.
Denoting the im2win tensor as $\mathcal{\hat I}$, the algorithm can be written as: 
\begin{equation}
\begin{aligned}
\mathcal{\hat I}(i, r, m, k H_{f}+u)=\mathcal{I}(i, r, m+u, n+v).
\end{aligned}
\end{equation}
subject to
\begin{equation}
\begin{aligned}
&m=0,1,..,H_{o}-1,  n=0,1,..,W_{o}-1, u=0,1,..,H_{f}-1,\\
&v~=0,1,..,W_{f}-1, i=0,1,..,N_{i}-1, r=0,1,..,C_{i}-1,\\
&k~=0,1,..,W_{i}-1. \nonumber
\end{aligned}
\end{equation}

{\small
\begin{algorithm}[t]
\SetAlgoNoLine
    \caption{Im2win Algorithm}
    \label{algorithm:im2win}
    \KwIn{Input tensor $\mathcal{I}$, Filter tensor $\mathcal{F}$, Stride $s$}
    \KwOut{Im2win tensor $\mathcal{\hat I}$}
    $H_o = (H_i-H_f)/s+1$ \label{algorithm:im2win:line:calculate_Ho}\\
    \uIf{$H_f>s$}{Allocate $\mathcal{\hat I}$ with $N_i \times H_o \times C_i \times H_f \times W_i$}\label{algorithm:im2win:line:allocate_Wf>s}
    \uElse{Allocate $\mathcal{\hat I}$ with $N_i \times C_i \times H_i \times W_i$}\label{algorithm:im2win:line:allocate_Wf<s}
    \For{$i=0$ \bf{to} $N_{i}-1$ \bf{in parallel}}{\label{algorithm:im2win:line:copy_Ni}
        \For{$r=0$ \bf{to} $C_{i}-1$}{\label{algorithm:im2win:line:copy_Ci}
            \For{$m=0$ \bf{to} $H_{o} -1$}{\label{algorithm:im2win:line:copy_Ho}
                \For{$k=0$ \bf{to} $W_{i}-1$}{\label{algorithm:im2win:line:copy_Wi}
                    \For{$u=0$ \bf{to} $H_{f}-1$}{\label{algorithm:im2win:line:copy_Hf}
                        $\mathcal{\hat I}[i][r][m][k*H_{f}+u]$ = $\mathcal{I}[i][r][m*s +u][k]$}}}}}\label{algorithm:im2win:line:copy}
\end{algorithm}
}

The im2win algorithm is described in~\Cref{algorithm:im2win}. 
At~\Cref{algorithm:im2win:line:calculate_Ho}, we first calculate the height of $\mathcal{O}$ ($\mathcal{O}$ is computed in the subsequent im2win-based convolution algorithm).
Next we decide the size/dimension of the im2win tensor $\mathcal{\hat I}$  from \Cref{algorithm:im2win:line:allocate_Wf>s} to~\Cref{algorithm:im2win:line:allocate_Wf<s}.
When $H_f>s$, the lower part of the dot product windows in the previous row and the upper part of the dot product windows in the current row have the same elements, therefore we allocate memory of size $N_i\times H_o\times C_i\times H_f\times W_i$ to $\mathcal{\hat I}$ (\Cref{algorithm:im2win:line:allocate_Wf>s}).
For example in~\Cref{fig:im2win_illustration} $H_f=2$ and $s=1$, therefore $H_f>s$. Note that in the input tensor, the elements in the second row of the first dot product window (the red box) and the elements of the first row of the fourth dot product window (the orange box) are the same (colored in gray). Similarly, the second row of the second window and first row of the fifth window are the same; the second row of the third window and the first row of the sixth window are the same.
When $H_f\leq s$, 
there are no common elements in the lower part of the dot product windows in the previous row and the upper part of the dot product windows in the current row. In this case, the size of $\mathcal{\hat I}$ is the same as that of $\mathcal{I}$. 
We allocate memory of size $N_i\times C_i\times H_i\times W_i$ to $\mathcal{\hat I}$ (\Cref{algorithm:im2win:line:allocate_Wf<s}).

Next, the elements of $\mathcal{I}$ are copied into $\mathcal{\hat I}$. Recall that the underlying storage of all data is an 1D array in memory. If the tensor is stored in the order of $NCHW$ structure, the distance of the memory addresses of two contiguous elements in the $W$ dimension is one and the distance of the memory addresses of two contiguous elements in the $H$ dimension is $W$. In hierarchical memory architecture,
because elements with consecutive memory addresses tend to be pulled up the memory hierarchy together.
The shorter the distance of the memory addresses of the elements is, the better the spatial locality is. Because of this, the access cost for $N, C, H, W$ dimensions decreases respectively.
Hence we prioritize the loop order (\Cref{algorithm:im2win:line:copy_Ni} -~\Cref{algorithm:im2win:line:copy_Hf}) based on the access cost of $\mathcal{\hat I}(N_i,~C_i,~H_o,~W_i \times H_f)$ when copying the elements from $\mathcal{I}$ into $\mathcal{\hat I}$ (\Cref{algorithm:im2win:line:copy}).

Now we compare the memory usage of the im2win tensor with that of the im2col matrix.
Suppose the size of $\mathcal{I}$ is $N_i \times C_i \times H_i \times W_i$, 
the size of $\mathcal{F}$ is $N_f \times C_f \times H_f \times W_f$, and the stride is $s$ (notice that $H_f \nless s$ and $W_f \nless s$).
When $H_f > s$, the size of the im2win tensor is $N_i\times C_i\times H_f\times W_i \times ((H_i-H_f)/s+1)$. When $H_f = s$, the size of the im2win tensor is equal to the size of $\mathcal{I}$. 
Additionally, the size of the im2win tensor has no correlation with the value of $W_f$.
When $H_f > s$ or $W_f > s$, the size of the im2col matrix is $C_i \times H_f \times W_f \times ((H_i-H_f)/s+1) \times ((W_i-W_f)/s+1)$, which is greater than the size of $\mathcal{I}$. When $H_f =W_f=s$, the size of the im2col matrix is equal to that of $\mathcal{I}$. Assume $H_f = W_f$,
the size difference of the im2col matrix and the im2win tensor is $\delta = H_f \times C_i \times(W_i-W_f)\times((H_i-H_f)/s+1)\times(W_f/s-1)$, where $H_o, H_f, 
C_i, W_i, W_f, s$ are all greater than 0 and $W_i \geq W_f$. 
When $H_f=W_f=s$, we get $\delta=0$, i.e., the im2win tensor and the im2col matrix has the same size of $\mathcal{I}$. When $H_f=W_f>s$, we get $\delta>0$, i.e., the size of the im2col matrix is greater than that of the im2win tensor. In summary, the size of the im2col matrix is greater than that of the im2win tensor no matter what.
Recall in~\Cref{fig:im2col_illustration} $s=1$, the im2col matrix has 36 elements, while in~\Cref{fig:im2win_illustration}, our im2win tensor has 24 elements. Our im2win tensor has $1/3$ less elements than the im2col matrix in addition to better data locality.
{\small
\begin{algorithm}[t]
\SetAlgoNoLine
    \caption{Basic im2win-based Convolution}
    \label{algorithm:im2win-based_convolution}
    \KwIn{Input tensor $\mathcal{I}$, Filter tensor $\mathcal{F}$, Stride $s$}
    \KwOut{Output tensor $\mathcal{O}$}
    $\mathcal{\hat I}[N_i][C_i][H_o][W_i \times H_f]$ = \textbf{Function} \textsc{im2win}($\mathcal{I}, \mathcal{F}, s$)\label{algorithm:im2win-based_convolution:im2win_transformation}
    \SetKwProg{func}{Function}{}{}
    
    \For{$i=0$ \bf{to} $N_{o}-1$}{
        \For{$j=0$ \bf{to} $C_{o}-1$}{
            \For{$m=0$ \bf{to} $H_{o}-1$}{
                \For{$n=0$ \bf{to} $W_{o}-1$}{
                    \For{$r=0$ \bf{to} $C_{f}-1$}{
                        \For{$u=0$ \bf{to} $H_{f}-1$}{
                            \For{$v=0$ \bf{to} $W_{f}-1$}{
                                $\mathcal{O}[i][j][m][n]$ += $\mathcal{\hat I}[i][r][m][n*s*W_f+v*H_f+u] \times \mathcal{F}[j][r][u][v]$}}}}}}} \label{algorithm:basic_im2win-based_conv:line:axpy}
 \end{algorithm}
}

\subsection{Our Im2win-based Convolution Algorithm}
We propose an im2win-based convolution, as shown in~\Cref{algorithm:im2win-based_convolution}.
The input tensor $\mathcal{I}$ is transformed into the im2win tensor $\mathcal{\hat I}$ at~\Cref{algorithm:im2win-based_convolution:im2win_transformation}. Next, the convolution is implemented as multiple nested for loop structure akin to direct convolution. In the innermost loop, the most expensive computation is the AXPY operation at~\Cref{algorithm:basic_im2win-based_conv:line:axpy}.~\Cref{algorithm:im2win-based_convolution} is the basic implementation before optimization. 
We propose a composition of optimizations making the best use of the im2win data transformation. 
These optimizations include
vectorization and FMA instructions, loop reordering, hoist and loop unrolling, register and cache blocking, and a parallelization strategy.

\noindent\textbf{Vectorization and FMA instructions.}
We vectorize the basic im2win-based convolution algorithm using the Single Instruction Multiple Data (SIMD) instruction set and the FMA instructions. 
Each FMA instruction operates on $N_{vec}$ scalar output elements simultaneously. In addition, there are a total of $N_{reg}$ logical registers that can be addressed, so the number of elements that the registers can retain is $N_{reg}N_{vec}$.
We used AVX2 instruction set in our implementation. This instruction set has 16 logical registers, each of which can store 256-bit of data.
The most expensive computation in our algorithm is the AXPY operation (\Cref{algorithm:basic_im2win-based_conv:line:axpy} in~\Cref{algorithm:im2win-based_convolution}), thus we implement it using the FMA instructions. 

\noindent\textbf{Loop reordering, hoist and loop unrolling.} 
\Cref{algorithm:im2win-based_convolution} has seven nested for loops. Each for loop
 is independent of the others, hence we can arbitrarily change the order of the loops without affecting the computational results. 
Because the access costs of different dimensions of a tensor are different,
to take advantage of the spatial locality of loading,
we should prioritize the computation of dimensions that are less expensive to access and that exhibit consecutive memory access. 
Hence we reorder the loop structure of~\Cref{algorithm:im2win-based_convolution} to the loop structure of~\Cref{algorithm:High_performance_in2win-based_convolution}.


In~\Cref{algorithm:im2win-based_convolution}, $\mathcal{\hat I}$ is the most expensive tensor to access, 
so we should minimize the number of access to it. We observe that 
$\mathcal{\hat I}$ is independent of the $C_o$ dimension of $\mathcal{O}$ and $\mathcal{F}$ (\Cref{algorithm:basic_im2win-based_conv:line:axpy} in~\Cref{algorithm:im2win-based_convolution}), therefore we hoist the elements of $\mathcal{\hat I}$ to keep them in registers until they are no longer used (\Cref{algorithm:High_performance_im2win:line:load_I_begin} -~\Cref{algorithm:High_performance_im2win:line:load_I_end} in~\Cref{algorithm:High_performance_in2win-based_convolution}). 
After loading the elements of $\mathcal{F}$ in the innermost loop (\Cref{algorithm:High_performance_im2win:line:load_F_end} in~\Cref{algorithm:High_performance_in2win-based_convolution}), the FMA operation performs AXPY operation (\Cref{algorithm:High_performance_im2win:line:FMA} in~\Cref{algorithm:High_performance_in2win-based_convolution}).

After determining the overall loop order of the algorithm, we consider the unrolling of individual loops. Loop unrolling reduces the number of cache/memory read, and also reduces the number of mis-branch predictions generated by the processor in each iteration.  
 We observe that the innermost loop in~\Cref{algorithm:im2win-based_convolution} has loop independence, hence we unroll (also called as flatten)~\cite{unrolling} the innermost loop  (\Cref{algorithm:High_performance_im2win:line:unloop_begin} -~\Cref{algorithm:High_performance_im2win:line:unloop_end} in~\Cref{algorithm:High_performance_in2win-based_convolution}) in our implementation. 
{
\small
\begin{algorithm}[t]
\SetAlgoNoLine
    \caption{High Performance im2win-based Convolution Algorithm}
    \label{algorithm:High_performance_in2win-based_convolution}
    \KwIn{Input tensor $\mathcal{I}$, Filter tensor $\mathcal{F}$, Stride $s$}
    \KwOut{Output tensor $\mathcal{O}$}
    $\mathcal{\hat I}[N_i][C_i][H_o][W_i \times H_f]$ = \textbf{Function} \textsc{im2win}($\mathcal{I},\mathcal{F},s$)
    \SetKwProg{func}{Function}{}{}
    
    \For{$jj=0$ \bf{to} $C_{o}/C_{o,b}-1$ \bf{in parallel}}{
        \For{$i=0$ \bf{to} $N_{o}-1$}{\label{algorithm:High_performance_im2win:line:outer_loop_begin}
            \For{$m=0$ \bf{to} $H_{o}-1$}{
                \For{$r=0$  \bf{to} $C_{f}-1$}{\label{algorithm:High_performance_im2win:line:outer_loop_end}      \For{$nn=0$ \bf{to} $W_{o}/W_{o,b}-1$}{
                    \For{$vv=0$ \bf{to} $W_{f}/W_{f,b}-1$}{     \sc{dot\_product}($jj,i,r,m,nn,vv,s$)}}}}}}
\func{\textsc{dot\_product}($jj,i,r,m,nn,vv,s$):}{
    \For{$n'=0$ \bf{to} $W_{o,b}-1$}{ \label{algorithm:High_performance_im2win:line:load_I_begin}\label{algorithm:High_performance_im2win:line:register_block_1}
        \For{$v'=0$ \bf{to} $W_{f,b}-1$}{\label{algorithm:High_performance_im2win:line:register_block_2}
                \For{$u=0$ \bf{to} $H_{f}/N_{vec}-1$}{\label{algorithm:High_performance_im2win:line:vector_load}
                $iu$ = $(nn*W_{o,b}+n')*s*W_f+(vv*W_{f,b}+v')*H_f+u$ \\
                    SIMD\_LOAD($\mathcal{\hat{I}}[i][r][m][iu:iu+N_{vec}]$)}}}\label{algorithm:High_performance_im2win:line:load_I_end}
    \For{$j'=0$ \bf{to} $C_{o,b}-1$}{\label{algorithm:High_performance_im2win:line:load_F_begin}\label{algorithm:High_performance_im2win:line:cache_block}
        SIMD\_LOAD($\mathcal{O}[i][jj*C_{o,b}+j'][m][nn]$) \\
        \For{$v'=0$ \bf{to} $W_{f,b}-1$}{
                \For{$u=0$ \bf{to} $H_{f}/N_{vec}-1$}{
                    SIMD\_LOAD($\mathcal{F}[jj*C_{o,b}+j'][r][vv*W_{f,b}+v'][u:u+N_{vec}]$) \label{algorithm:High_performance_im2win:line:load_F_end}\label{algorithm:High_performance_im2win:line:unloop_begin}\\ 
                    FMA($\mathcal{\hat I},\mathcal{F},\mathcal{O}[i][jj*C_{o,b}+j'][m][nn]$)\label{algorithm:High_performance_im2win:line:unloop_end}\label{algorithm:High_performance_im2win:line:FMA}}}
                    SIMD\_STORE($\mathcal{O}[i][jj*C_{o,b}+j'][m][nn]$)
                    }}
\end{algorithm}
}

\noindent\textbf{Register and cache blocking.}
The minimum number of output elements $\theta$ required to maintain the peak performance of the SIMD architecture system is limited by the maximum number of elements that can be kept in all vector registers~\cite{analyrical_modeling_blis}, i.e., $\theta \leq N_{reg} N_{vec}$.
To achieve maximum performance of our algorithm, we want to keep as many elements of tensors as possible in registers and minimize the number of SIMD load. Therefore, we adopt register blocking~\cite{tiling} to load elements of several dot product windows into registers (dynamically calculated based on $W_f$, $H_f$ and $N_{vec}$) at~\Cref{algorithm:High_performance_im2win:line:load_I_end} and~\Cref{algorithm:High_performance_im2win:line:load_F_end} respectively in~\Cref{algorithm:High_performance_in2win-based_convolution}.
 We apply register blocking to $W_o$ and $W_f$ dimensions at~\Cref{algorithm:High_performance_im2win:line:register_block_1} -~\Cref{algorithm:High_performance_im2win:line:register_block_2} in~\Cref{algorithm:High_performance_in2win-based_convolution} (omitted due to space limit), followed by a SIMD\_LOAD in the filter height dimension $H_f$ at~\Cref{algorithm:High_performance_im2win:line:vector_load} in~\Cref{algorithm:High_performance_in2win-based_convolution}. 


At the cache level, we can partition the input data to fit the cache size to minimize cache misses.
Since our operation process is to continuously iterate a new filter window in the $C_o$ dimension of output  and perform dot product with $\mathcal{\hat I}$, thus improving the data reusability of the elements in $\mathcal{\hat I}$. We choose to divide the iterative output channel dimension $C_o$ into smaller partitions to fit the memory to the next level of the hierarchy (\Cref{algorithm:High_performance_im2win:line:cache_block} in~\Cref{algorithm:High_performance_in2win-based_convolution}).
 
\noindent\textbf{Parallelization strategy.}
We observe in our algorithm that all the output elements are independent of each other and are embarrassingly parallelizable~\cite{art}. Since the output is a four-dimensional tensor $\mathcal{O}(N_i \times C_o \times H_o \times W_o)$, this means that we can pick any one of four dimensions for parallelization. Our im2win-based convolution implementation extracts parallelism in the output channel $C_o$ dimension. Each thread is assigned a block of output elements of size $H_o\times W_o\times C_o/t$, where $t$ is the number of threads used.

\begin{figure*}[ht] 
\centering
\subfigure{
\includegraphics[scale=0.42]{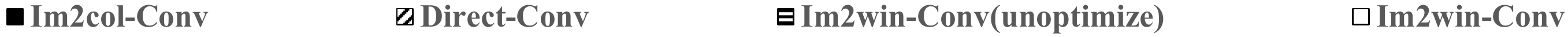}
    } \\
    \subfigure[Normalized Runtime]{\label{fig:performance:subfig:NormalizedRunTime}
        \includegraphics[scale=0.42]{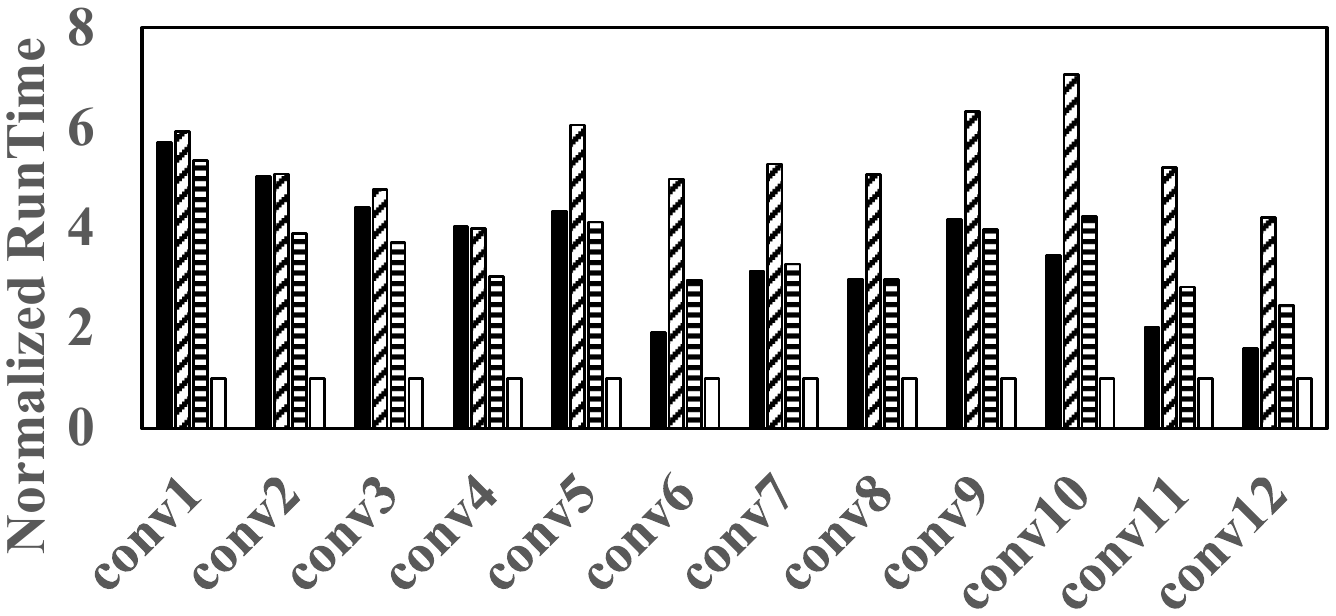}
    }
    \subfigure[GFLOPS]{\label{fig:performance:subfig:GFLOPS}
        \includegraphics[scale=0.42]{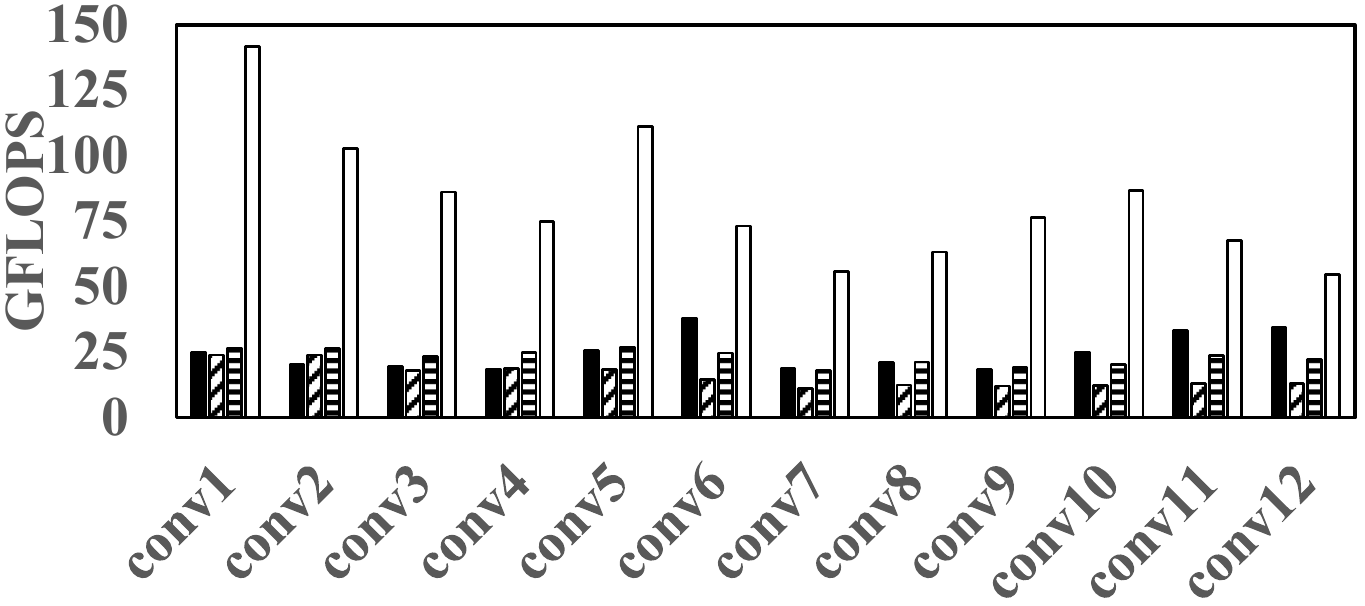}
    }\subfigure[Normalized Speedup]{\label{fig:performance:subfig:NormalizedGFLOPS}
        \includegraphics[scale=0.42]{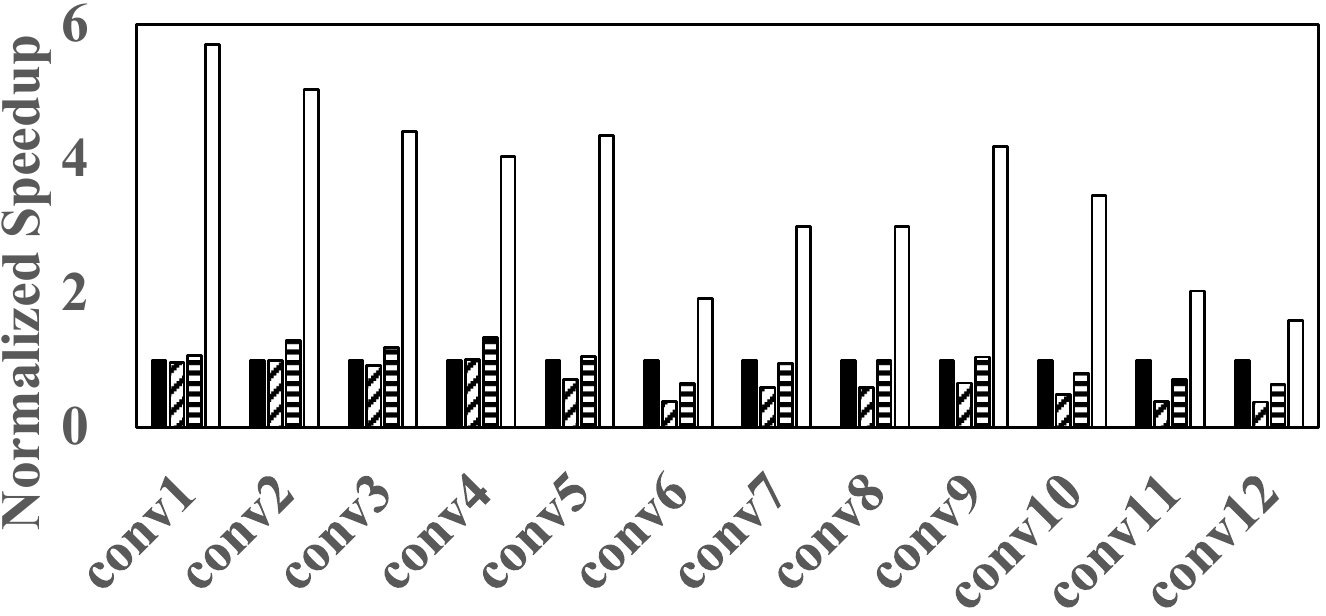}
    }
    \caption{\small Performance comparison of our convolution algorithms with PyTorch's im2col-based convolution and direct convolution. The runtime results are normalized to our convolution, and the performance results in GFLOPS are normalized to im2col-based convolution.
    Among twelve convolution layers, our convolution has the fastest runtime, while PyTorch's runtime is average to 3.6$\times$ slower and the direct convolution's runtime is average to 5.3$\times$ slower than us. Our convolution has up to about 3.6$\times$ GFLOPS than that of Pytorch and 5.3$\times$ that of direct convolution, respectively.}
    \label{fig:performance}
\end{figure*}

\section{Experimental Results}
In this section, we compare the performance results and memory usages of our im2win-based convolution algorithms with Pytorch's im2col-based algorithm and a direct convolution implementation.  
\subsection{Experimental Setup}
\noindent\textbf{Platform.} We run the experiments on two Intel Xeon Silver 4214 processors at 2.20GHz (each has 12 cores). 
Each CPU has 32KB L1 cache, 1MB L2 cache, and 16.5MB L3 cache. \\
\noindent\textbf{Software.} We implement our im2win-based convolution algorithm in C++ and parallelize it using OpenMP 4.5. We
use the tensor data structure of PyTorch 1.10.0a0~\cite{pytorchlib}, and the element in each tensor is a single-precision number.
We compare our convolution algorithms with PyTorch's im2col-based convolution and direct convolution in terms of performance and memory usage. The im2col-based convolution algorithm in PyTorch is parallelized by OpenMP.
The direct convolution code implemented in this work~\cite{directconvolutions} is not open-sourced and uses a special data layout, so we implemented a naive direct convolution in C++ using nested loops and parallelized with OpenMP. 
We compile all the algorithms with GCC 7.5 and ``-Ofast -march=native'' compilation options.

\noindent\textbf{Benchmarks.} We want to cover the majority of the convolutional layers used in commonly used DNNs. However, 
it is impossible if we benchmark with one neural network model with just one filter size for all convolutional layers, for example, all filter tensors in VGG-16~\cite{VGG-16} are $3\times3$, and ResNet-50~\cite{resnet} contains only three different filters in sizes.
Hence for our experimental evaluation, we select an state-of-the-art DNN benchmark~\cite{mec}, which includes twelve unique convolution layers, Conv1-Conv12 (shown in~\Cref{table:benchmarks}).

\begin{table}[htbp]
\caption{\small Twelve convolution layers of the DNN benchmarks.}
\label{table:benchmarks}
\begin{center}
\resizebox{\linewidth}{!}{
\begin{tabular}{cccc}
\toprule
\textbf{NAME}&\textbf{INPUT}&\textbf{FILTER, STRIDE}&\textbf{OUTPUT} \\
& $C_{i} \times H_{i} \times W_{i}$ & $C_{o} \times H_{f} \times W_{f}, s_{h}(s_{w})$&$C_{o} \times H_{o} \times W_{o}$\\
\midrule
$\textbf{Conv1}$ & $3\times227\times227$& $96\times11\times11, 4$&$96\times55\times55$\\
$\textbf{Conv2}$ & $3\times231\times231$& $96\times11\times11, 4$&$96\times56\times56$\\
$\textbf{Conv3}$ & $3\times227\times227$& $64\times7\times7, 2$&$64\times111\times111$\\
$\textbf{Conv4}$ & $64\times224\times224$& $64\times7\times7, 2$&$64\times109\times109$\\
$\textbf{Conv5}$ & $96\times24\times24$& $256\times5\times5, 1$&$256\times20\times20$\\
$\textbf{Conv6}$ & $256\times12\times12$& $512\times3\times3, 1$&$512\times10\times10$\\
$\textbf{Conv7}$ & $3\times224\times224$& $64\times3\times3, 1$&$64\times222\times222$\\
$\textbf{Conv8}$ & $64\times112\times112$& $128\times3\times3, 1$&$128\times110\times110$\\
$\textbf{Conv9}$ & $64\times56\times56$& $64\times3\times3, 1$&$64\times54\times54$\\
$\textbf{Conv10}$ & $128\times28\times28$& $128\times3\times3, 1$&$128\times26\times26$\\
$\textbf{Conv11}$ & $256\times14\times14$& $256\times3\times3, 1$&$256\times12\times12$\\
$\textbf{Conv12}$ & $512\times7\times7$& $512\times3\times3, 1$&$512\times5\times5$\\
\bottomrule
\end{tabular}}
\end{center}
\end{table}
\subsection{Performance}
In the following experiments, we use the wall-clock time in the standard C++ library to measure the runtime of different algorithms. We run each algorithm five times and record the best runtime among five runs. The batch size of each convolution layer input data is 128.

\Cref{fig:performance:subfig:NormalizedRunTime} shows the normalized runtime of the different convolution algorithms. The runtime results are normalized to our optimized im2win convolution algorithm. Overall, our algorithm has the shortest runtime among all twelve different convolutional layers. Pytorch is up to 5.7$\times$ slower and the direct convolution is up to 7$\times$ slower than our algorithm, respectively.
Shown in~\Cref{fig:performance:subfig:GFLOPS}, We measure 
the floating-point computational performance of different convolutional algorithms, and normalize the results to our algorithm as~\Cref{fig:performance:subfig:NormalizedGFLOPS}. Our convolution has at most 6$\times$ GFLOPS than PyTorch, and at least 2$\times$ GFLOPS than PyTorch. Our convolution has a maximum of about 7$\times$ GFLOPS and a minimum of about 4$\times$ GFLOPS than direct convolution.

 

\subsection{Memory Usage}
\begin{figure}[htp]
\centering
\centerline{\includegraphics[scale=0.45]{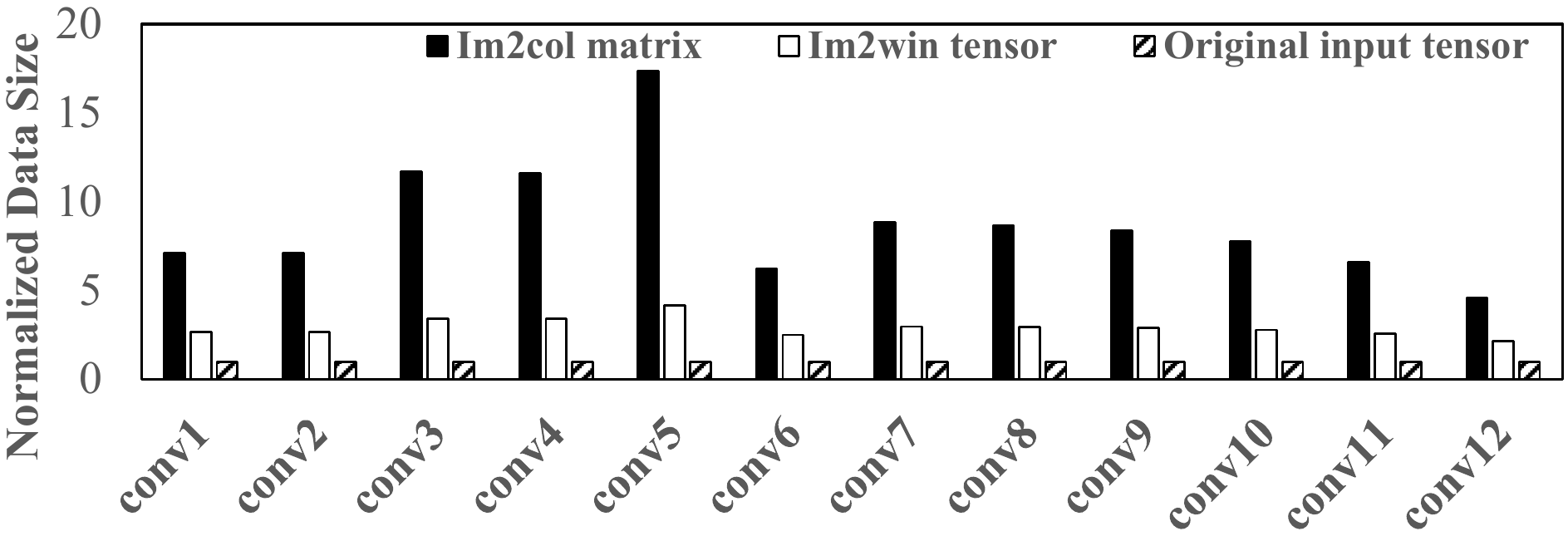}}
\caption{\small Normalized data sizes of the im2col matrix, the im2win tensor and the original input tensor of twelve convolution layers. We normalize all the data sizes to the original input tensor.}
\label{fig:DataSize}
\end{figure}

\begin{figure}[htp]
\centering
\centerline{\includegraphics[scale=0.45]{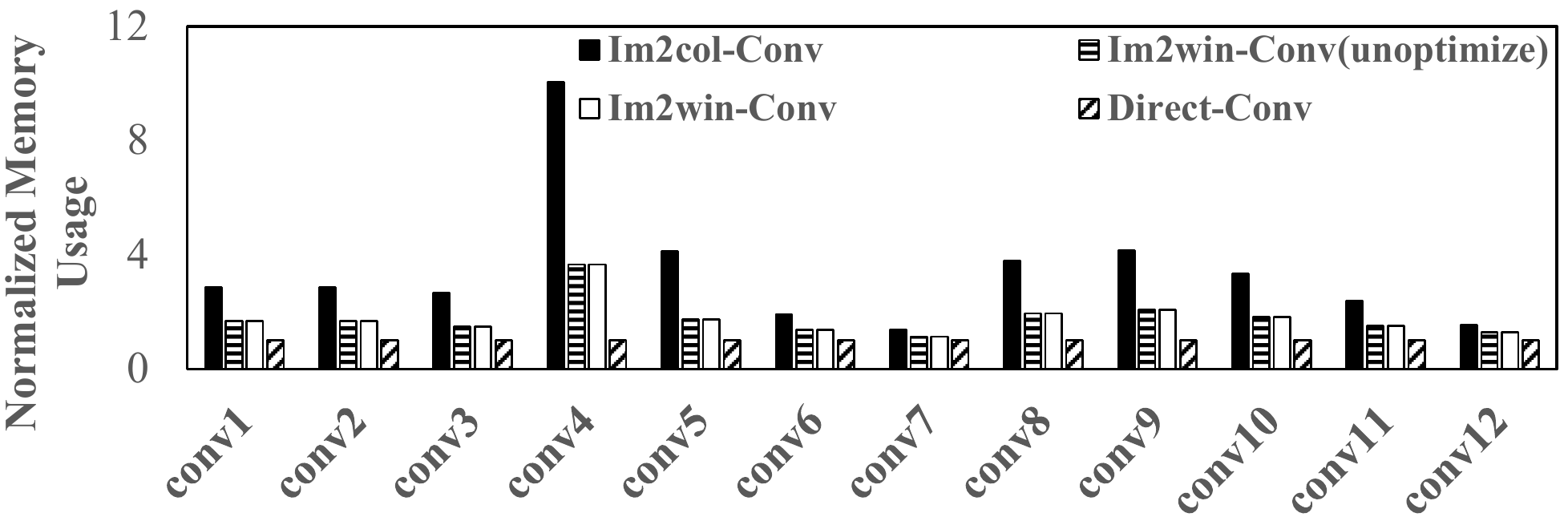}}
\caption{\small Normalized peak memory usage of all convolution algorithms on twelve convolution layers. Since direct convolution has no additional memory overhead of the data transformation, we normalize all the memory usage results to direct convolution.}
\label{fig:MemoryOverhead}
\end{figure}

We show the data sizes of the im2col matrix, the im2win tensor and the original input tensor of twelve convolution layers as~\Cref{fig:DataSize} and the peak memory usage results of all convolution algorithms as~\Cref{fig:MemoryOverhead}. On average, the im2col matrices are 2.93$\times$ larger than our im2win tensors for all twelve different convolutional layers. 
The im2col-based convolution uses up to 2.75$\times$ more memory than our im2win-based convolution. Our im2win-based convolution reduces memory overhead by average to 41.6\% compared to the im2col-based convolution algorithm. The results show that our convolution algorithm can more effectively reduce the memory footprint compared to the im2col-based convolution algorithm.




\section{Conclusion}
We proposed a memory-efficient im2win algorithm. Different from the classic im2col algorithm, im2win refactorizes a row of square or rectangle dot product windows of the input image and flattens unique elements within these windows into a row in the output tensor, which enables consecutive memory access and data reuse, and thus greatly reduces the memory overhead.
In addition, we proposed a high-performance im2win-based convolution algorithm, which implements a nested loop structure similar to direct convolution to improve data reusability and parallel performance. We tested on a benchmark containing most of the convolution layers in a neural network model. Experimental results show that our algorithm reduces memory overhead by average to 41.6\% compared to PyTorch's im2col-based convolution algorithm, and achieves speedups in performance of average to 3.6$\times$ and 5.3$\times$ compared to PyTorch's im2col-based convolution algorithm and direct convolution algorithm, respectively.

\section*{Acknowledgment} 
This work was supported by National Natural Science Foundation of China (Grant No. 62162045), Key Research and Development Program of Jiangxi (Program No. 20192BBE50073), and Technology Innovation Guidance Program Project of Jiangxi Province (Special Project of Technology Cooperation) (Grant No. 20212BDH81003).

\bibliographystyle{IEEEtran}
\bibliography{bibliography.bib}

\end{document}